%% file: 0-main.tex
\pdfoutput=1

\documentclass[letterpaper, 10 pt, conference]{ieeeconf}  

\IEEEoverridecommandlockouts                              

\usepackage[backend=bibtex,
            hyperref=true,
            url=false,
            isbn=false,
            doi=false,
            backref=false,
            style=ieee,
            natbib=true,
            mincitenames=1,
            maxcitenames=1,
            citestyle=numeric-comp,
            sorting=nyt,
            block=none]{biblatex}

\input{0-preamble.tex}


\title{\LARGE \bf
Safe Self-Supervised Learning in Real of 
\\Visuo-Tactile Feedback Policies for Industrial Insertion
}

\author{Letian Fu$^{1,\dagger}$, Huang Huang$^{1}$, Lars Berscheid$^{3}$, Hui Li$^{2}$, Ken Goldberg$^{1}$, Sachin Chitta$^{2}$%
\thanks{$^{1}$The AUTOLab at UC Berkeley, $^{2}$Autodesk Research, $^{3}$Karlsruhe Institute of Technology \newline\indent$^\dagger$Work done as an intern at Autodesk}
}

\begin{document}

\maketitle

\begin{abstract}
Industrial insertion tasks are often performed repetitively with parts that are subject to tight tolerances and prone to breakage. Learning an industrial insertion policy in real is challenging as the collision between the parts and the environment can cause slippage or breakage of the part. In this paper, we present a safe self-supervised method to learn a visuo-tactile insertion policy that is robust to grasp pose variations. The method reduces human input and collisions between the part and the receptacle. The method divides the insertion task into two phases. In the first {\em align} phase, a tactile-based grasp pose estimation model is learned to align the insertion part with the receptacle. In the second {\em insert} phase, a vision-based policy is learned to guide the part into the receptacle. The robot uses force-torque sensing to achieve a safe self-supervised data collection pipeline. Physical experiments on the USB insertion task from the NIST Assembly Taskboard suggest that the resulting policies can achieve 45/45 insertion successes on 45 different initial grasp poses, improving on two baselines: (1) a behavior cloning agent trained on 50 human insertion demonstrations (1/45) and (2) an online RL policy (TD3) trained in real (0/45). \end{abstract}

\input{1-introduction.tex}
\input{2-related-work.tex}
\input{3-problem-statement.tex}
\input{4-methods.tex}
\input{5-experiment-and-result.tex}
\input{6-ablation.tex}
\input{7-summary.tex}


\renewcommand*{\bibfont}{\footnotesize}
\printbibliography 
\clearpage



\end{document}

%% file: 0-preamble.tex
\usepackage{graphics}
\usepackage[pdftex]{graphicx}
\usepackage{xcolor}
\usepackage{wrapfig}
\DeclareGraphicsExtensions{.pdf,.png,.jpg}
\usepackage{epsfig}
\usepackage[font={small}]{caption}
\usepackage{subcaption}
\usepackage[rightcaption]{sidecap}
\usepackage{pbox}
\usepackage{makecell}

\usepackage{bigstrut}
\usepackage{mathtools}
\usepackage{amssymb,amsmath} 
\usepackage{gensymb} 
\usepackage{nicefrac}       
\numberwithin{equation}{section} 

\DeclareMathAlphabet{\mathcal}{OMS}{lmsy}{m}{n}
\DeclareSymbolFont{largesymbols}{OMX}{cmex}{m}{n}
\usepackage{textcomp} 
\usepackage[linesnumbered, ruled,vlined,noend]{algorithm2e}

\usepackage{array} 
\usepackage{tabularx}
\usepackage{multirow}
\usepackage{multicol}
\usepackage{booktabs}
\usepackage{tabulary}

\usepackage[utf8]{inputenc}
\usepackage{bm}
\usepackage{xspace}
\usepackage{flushend}
\usepackage{balance} 
\usepackage{csquotes}
\usepackage{makeidx}
\usepackage{blindtext}

\usepackage{autolabtools}

\let\labelindent\relax
\usepackage{enumitem}

\usepackage{tikz}
\usetikzlibrary{backgrounds, positioning, calc, shapes.misc}

\usepackage{ragged2e}
\usepackage{soul} 
\usepackage{subfiles} 

\usepackage[protrusion=true,expansion=true]{microtype}
\usepackage{url}
\makeatletter
\g@addto@macro{\UrlBreaks}{\UrlOrds}
\makeatother

\usepackage{hyperref}
\hypersetup{
    colorlinks=true,
    linkcolor=black,
    citecolor=black,
    filecolor=cyan,
    urlcolor=black
}

\usepackage[separate-uncertainty=true]{siunitx}

\renewcommand{\baselinestretch}{1.0}
\usepackage{amsmath}

\DeclareMathOperator*{\argmin}{arg\,min}

\newcommand{\norm}[1]{\left\lVert#1\right\rVert}


%% file: 1-introduction.tex
\section{Introduction}
Industrial assembly~\cite{mckee1985automatic} is a precise manipulation task requiring contact between parts. Part feeding, peg insertion and object reorientation (three sub-tasks of industrial assembly) have been extensively studied~\cite{lozano1984automatic, lozano1986motion, goldberg1993orienting, natarajan1989some, mckee1985automatic}. Early work considers the mechanical design aspect~\cite{natarajan1989some,lozano1986motion} and motion planning aspect~\cite{lozano1984automatic, goldberg1993orienting, qiao1995fine}. Through Computer Aided Design (CAD), the order of part assembly can be predetermined in simulation with precise pose information~\cite{de1989correct}, allowing robots to plan the necessary actions to assemble the design~\cite{koga2022cad}. Learning-based approaches recently have shown promise on industrial insertion tasks~\cite{Wen2022once} on the NIST taskboard~\cite{NIST}, a standard benchmark that represents common industrial insertion tasks with parts that have complex geometries~\cite{Lian2021benchmark}. However, applying learning-based methods for industrial insertion remains challenging due to the requirement for frequent human inputs during learning~\cite{luo2021robust} or high-precision sensors for collecting training data~\cite{Wen2022once}. 
There is also a need for a safe training and data collection method for learning insertion tasks since parts are prone to breakage.

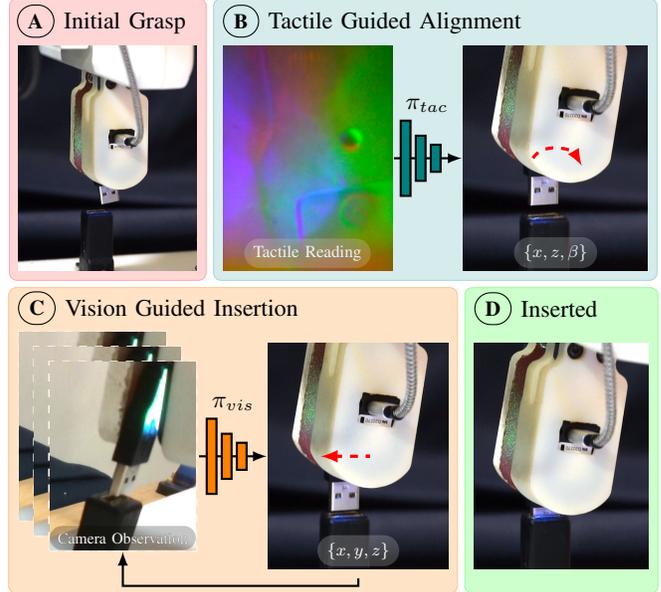
\begin{figure}[t!]
    \centering
    \input{figures/fig-1-tikz/fig.tex}
    \vspace{-0.1cm}
    \caption{Overview of the learned two-phase insertion policy: the red arrows indicating the robot actions given by the policies. (A) The robot grasps the part at an initial pose. (B) The tactile guided policy $\pi_{tac}$ estimates the grasp pose using the tactile image and aligns the z-axis of the part with the insertion axis. (C) A vision guided policy $\pi_{vis}$ is used to insert the part. (D) The part is inserted successfully into the receptacle.}
    \label{fig:splash}
\end{figure}
Another challenge in an industrial insertion task is that the precise grasp pose is often unknown due to variations in kitting and feeding of parts as they arrive for assembly. As the grasped part is often occluded by the gripper visually, grasp pose estimation is better achieved when using tactile sensing~\cite{okumura2022tactile}. While recent work has shown improvement in simulation accuracy for industrial insertion~\cite{narang2022factory} and successes in Sim2Real transfer for tactile-based insertion tasks~\cite{kelestemur2022tactilepose, Wang2022TACTO}, the simulation of soft contacts between tactile sensors and objects with complex geometries remains an open problem~\cite{wang2021elastic}, and often can not transfer to real because the object models are not publicly available. In this work, we present a novel method to safely learn visuo-tactile feedback policies in real for industrial insertion tasks under grasp pose uncertainties, with inexpensive off-the-shelf sensors. Our approach draws on tactile and visual feedback to deal with the grasp pose uncertainty and force-torque sensing for a self-supervised training procedure that is {\em safe}, minimizing damage during the training phase. We divide the insertion task into two phases (as shown in Fig.~\ref{fig:splash}): 
\begin{itemize}
    \item An initial {\em Align} phase where a tactile-based policy $\pi_{tac}$ estimates the grasp pose. The robot reorients and aligns the part with the insertion axis of the receptacle.
    \item A second {\em Insert} phase where an RGB image-based policy $\pi_{vis}$ guides the robot to insert the part. 
\end{itemize}

A significant challenge in learning tactile feedback policies in real for industrial insertion is the frequent slippage of the part that occurs due to collision with the environment and the smooth surface of the tactile sensor gel pad. This makes RL methods difficult to succeed without human intervention or an automatic reset mechanism to detect and correct slippage. In this work, we develop a self-supervised data collection pipeline that avoids collision between the part and its environment, by recognizing that the insertion operation is reversible only from certain target insertion poses – i.e. starting from such poses, the part can be repeatedly unplugged from and inserted into the receptacle. Prior to data collection, a human free-drives the robot to provide one approximate target pose where the part is inserted. The robot refines this target pose to find such a reversible pose by minimizing the grasping force-torque, which helps minimize collisions during data collection, resulting in a safer training process that is unlikely to damage the insertion part and receptacle. 

This paper contributes:
\begin{enumerate}
    \item A safe self-supervised data collection pipeline with force-torque sensing in real for insertion, designed to minimize contact force for data collection.
    \item A two-phase policy learned from the collected data including a tactile-based alignment policy for orienting the part and an RGB image-based insertion policy;
    \item Experimental results suggest that the policy achieves 45/45 successes on USB connector insertion, outperforming two baseline methods (1/45 and 0/45).
\end{enumerate}

%% file: figures/fig-1-tikz/fig.tex
\begin{tikzpicture}[font=\small]
	\node[] (i1) {\includegraphics[height=3cm, trim=0 275 0 0, clip]{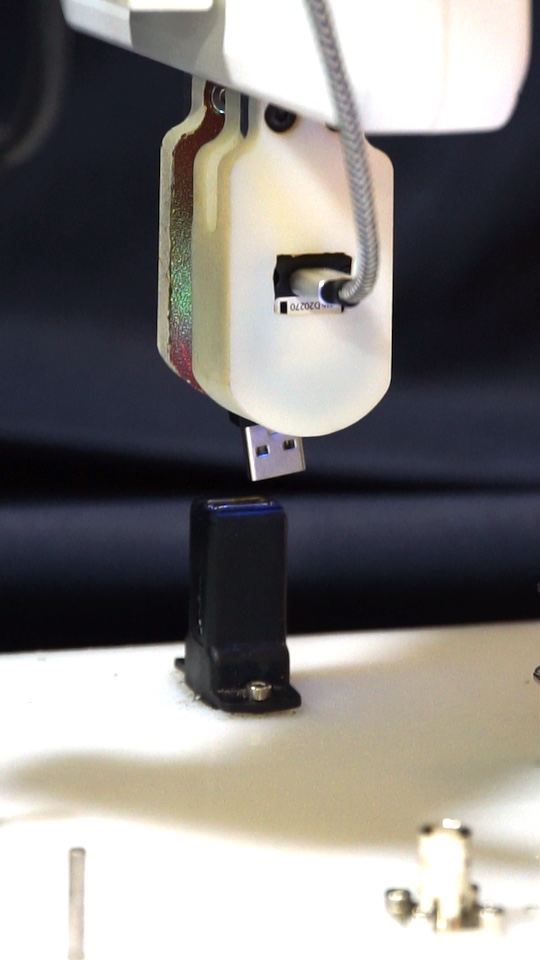}};
	
	\node[right=1mm of i1] (i2) {\includegraphics[height=3cm]{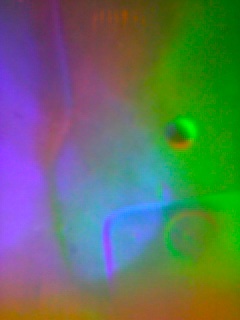}};
	\node[right=7mm of i2] (i3) {\includegraphics[height=3cm, trim=81 350 100 170, clip]{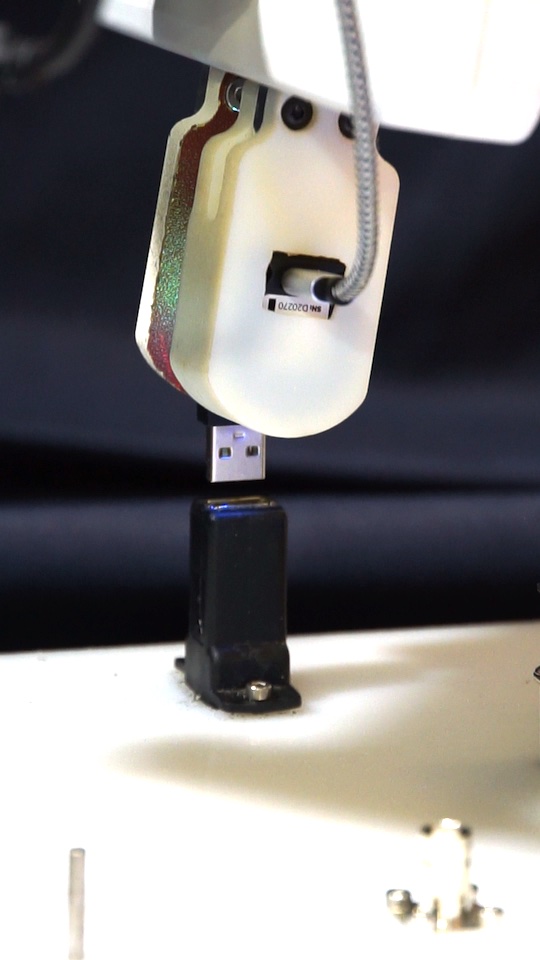}};

	\node[below=-5.6mm of i2, rounded rectangle, fill=black, fill opacity=0.2, text opacity=1.0, font=\footnotesize, teal!8, scale=0.8] {Tactile Reading};
	
	\coordinate (mid2) at ([xshift=-0.6mm]$(i2)! 0.5!(i3)$);
	\draw[thick, -latex] ([xshift=-1mm]i2.east) -- ([xshift=1mm]i3.west);
	\draw[thick, fill=teal] (mid2)++(-0.35, -0.5) rectangle ++(0.13, 1.0);
	\draw[thick, fill=teal] (mid2)++(-0.15, -0.3) rectangle ++(0.13, 0.6);
	\draw[thick, fill=teal] (mid2)++(0.05, -0.18) rectangle ++(0.13, 0.36);
	\node[yshift=7mm] at (mid2) {$\pi_{tac}$};
	
	\draw[dashed, very thick, red, -latex] (i3)++(-0.3, 0.0) to[in=120, out=60] ++(0.65, -0.1);
	
	\node[below=-5.8mm of i3, rounded rectangle, fill=black, fill opacity=0.2, text opacity=1.0, font=\footnotesize, teal!8, scale=0.8] {$\lbrace x, z, \beta \rbrace$};
	
	\node[xshift=1mm, yshift=2mm, anchor=west, rounded rectangle, draw=black] (n1) at (i1.north west) {\footnotesize \textbf{A}};
	\node[right=0mm of n1, yshift=-0.25mm] {Initial Grasp};
	
	\node[xshift=1.25mm, yshift=2mm, anchor=west, rounded rectangle, draw=black] (n1) at (i2.north west) {\footnotesize \textbf{B}};
	\node[right=0mm of n1, yshift=-0.25mm] {Tactile Guided Alignment};

	\node[below=6.7mm of i1, xshift=-2.0mm, draw=orange!10, dashed, very thick, inner sep=0] (i4a) {\includegraphics[height=2.5cm, trim=420 180 10 30, clip]{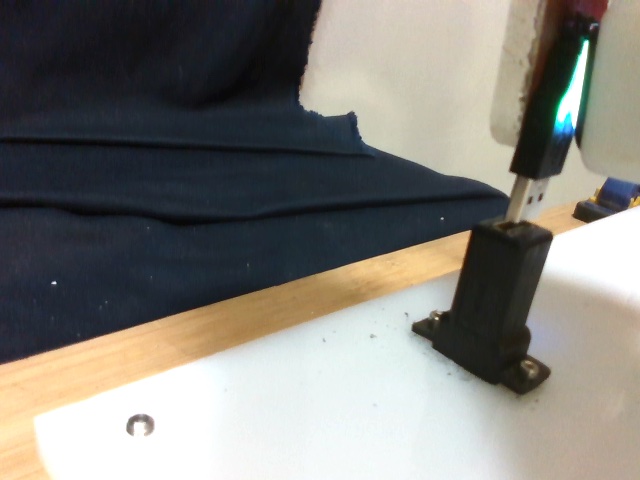}};
	\node[xshift=2mm, yshift=-2mm, draw=orange!10, very thick, dashed, inner sep=0] (i4b) at (i4a) {\includegraphics[height=2.5cm, trim=420 180 10 30, clip]{figures/fig-1-tikz/4-cam-obs.jpg}};
	\node[xshift=2mm, yshift=-2mm, draw=orange!10, very thick, dashed, inner sep=0] (i4) at (i4b) {\includegraphics[height=2.5cm, trim=420 180 10 30, clip]{figures/fig-1-tikz/4-cam-obs.jpg}};
	
	\node[right=8.2mm of i4] (i5) {\includegraphics[height=3cm, trim=90 350 100 170, clip]{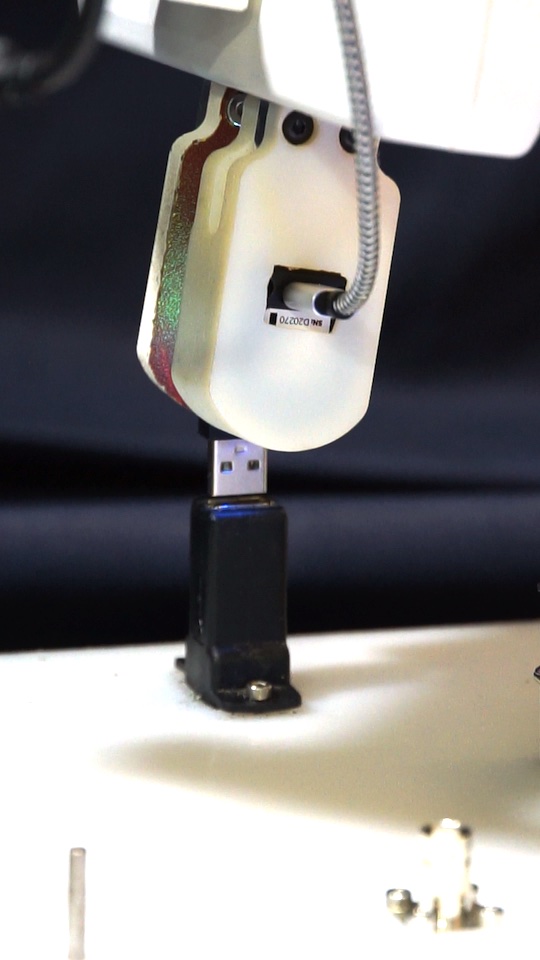}};
	\node[right=1mm of i5] (i6) {\includegraphics[height=3cm, trim=100 350 100 170, clip]{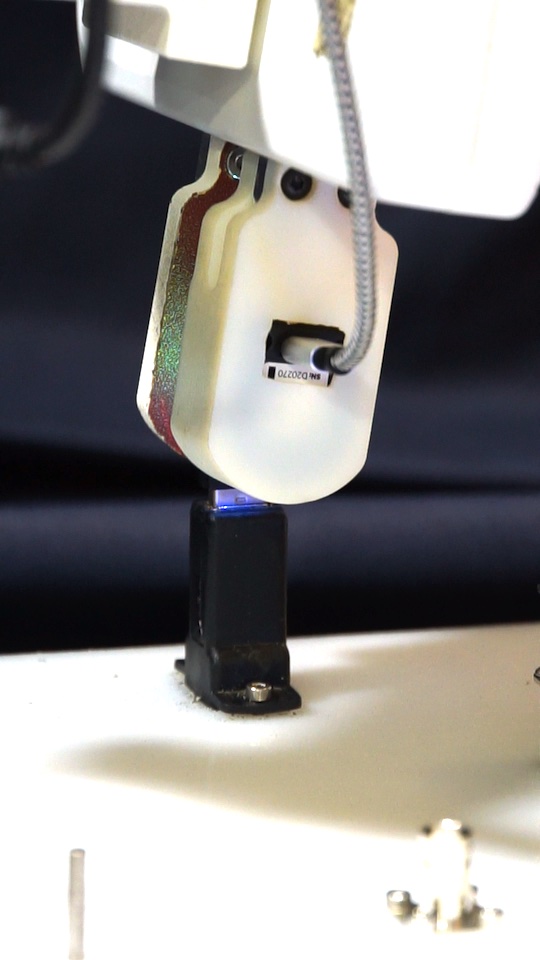}};
	\node[below=-3.5mm of i4, rounded rectangle, fill=black, fill opacity=0.23, text opacity=1.0, font=\footnotesize, color=black, scale=0.74] {Camera Observation};
	\node[below=-3.5mm of i4, rounded rectangle, fill=orange!50, fill opacity=0.0, text opacity=1.0, font=\footnotesize, orange!2, scale=0.74] {Camera Observation};
	
	\coordinate (mid3) at ([xshift=-1mm]$(i4)! 0.5!(i5)$);
	\draw[thick, -latex] ([xshift=0.2mm]i4.east) -- ([xshift=1mm]i5.west);
	\draw[thick, fill=orange] (mid3)++(-0.35, -0.5) rectangle ++(0.13, 1.0);
	\draw[thick, fill=orange] (mid3)++(-0.15, -0.3) rectangle ++(0.13, 0.6);
	\draw[thick, fill=orange] (mid3)++(0.05, -0.18) rectangle ++(0.13, 0.36);
	\node[yshift=7mm] at (mid3) {$\pi_{vis}$};
	
    \draw[dashed, very thick, red, -latex] (i5)++(0.1, 0.) to[in=0, out=0] ++(-0.6, 0.);
	\node[below=-5.8mm of i5, rounded rectangle, fill=black, fill opacity=0.2, text opacity=1.0, font=\footnotesize, white!2, scale=0.8] {$\lbrace x, y, z \rbrace$};
	
	\draw[thick, -latex] (i5.south) |- ++(0, -0.1)  -| (i4.south);
	
	\node[yshift=3mm, anchor=west, rounded rectangle, draw=black] (n1) at (i4a.north west) {\footnotesize \textbf{C}};
	\node[right=0mm of n1] {Vision Guided Insertion};
	
	\node[xshift=1.25mm, yshift=3.4mm, anchor=west, rounded rectangle, draw=black] (n1) at (i6.north west) {\footnotesize \textbf{D}};
	\node[right=0mm of n1] {Inserted};

	\begin{scope}[on background layer]
		\draw[draw=red!32, thin, fill=red!16, rounded corners=1mm] (i1.south west) rectangle ([yshift=5mm]i1.north east);
		
		\draw[draw=teal!32, thin, fill=teal!16, rounded corners=1mm] (i2.south west) rectangle ([yshift=5mm]i3.north east);
		
		\draw[draw=orange!44, thin, fill=orange!22, rounded corners=1mm] ([yshift=-9.5mm, xshift=-1.25mm]i4a.south west) rectangle ([yshift=6.2mm]i5.north east);
		
		\draw[draw=green!40, thin, fill=green!20, rounded corners=1mm] ([yshift=-2.0mm]i6.south west) rectangle ([yshift=6.2mm]i6.north east);
	\end{scope}
\end{tikzpicture}

%% file: 2-related-work.tex
\section{Related Work}
\label{sec:related_works}
Industrial insertion has been central in robotics for 50 years. It is challenging due to occlusions brought by the robot gripper, grasp uncertainty from the process of acquiring the part and its collision with the environment, the fragility of the parts, and the precision required in controlling the robot for insertion. Early work approached this problem using CAD information to infer desired assembly sequences~\cite{de1989correct} and generating designs of part feeders based on object geometry~\cite{natarajan1989some}. Other work approached the problem from an algorithmic design perspective, with a focus on developing motion planning strategies for peg insertion~\cite{lozano1986motion,qiao1995fine}. 

Recently, learning-based methods have shown success on this task. This includes learning insertion policies with a physical robot via Sim2Real transfer~\cite{johannink2019residual}, online adaptation with meta-learning~\cite{schoettler2020meta, zhao2022offline}, reinforcement learning~\cite{schoettler2020deep, luo2021robust}, self-supervised data collection with impedance control~\cite{Spector2021insertionnet}, accurate state estimation~\cite{Wen2022once}, or decomposing the insertion algorithm into a residual policy that relies on conventional feedback control~\cite{johannink2019residual}. These approaches assume that the parts are grasped with a fixed pose. To overcome this assumption, \citet{Wen2022once} perform accurate pose estimation and motion tracking with a high-precision depth camera and use a behavioral cloning algorithm to insert the part. \citet{Spector2021insertionnet, Spector2022insertionnet2} require contact between the part and the environment to occur during data collection, a process that is expensive and often impractical for fragile parts. In comparison, we use inexpensive tactile sensors and a safe self-supervised data collection procedure that does not require such contact. 

Grasped parts are often visually occluded by the gripper. Tactile feedback can be an alternative sensing modality for grasp pose estimation.  Recent work uses tactile images from vision-based tactile sensors such as GelSight~\cite{Yuan2017gelsight} and DIGIT~\cite{lambeta2020digit} to estimate the pose of grasped objects. \citet{Li2014gelsight} use Gelsight sensors, BRISK features and RANSAC to estimate grasp pose. Gelsight produces high-quality 3D tactile images and can determine depth imprint, which improves feature detection by isolating the object from the background. DIGIT, a more affordable tactile sensor, provides a 2D RGB image but not the light incident direction (to generate the depth image). \citet{kelestemur2022tactilepose} generates tactile image data in simulation for pose estimation of bottle caps but simulating contact and physical interaction between tactile sensors and objects with more intricate geometry is still challenging~\cite{wang2021elastic}. In this work, we collect a dataset of tactile images in real for the USB connector with different grasp poses to train a tactile based policy for grasp pose estimation. 

Most prior work on tight tolerance insertion tasks~\cite{Wen2022once, Florence2021ibc, Li2014gelsight, Fan2019framework} leverages a single modality, such as vision, tactile, or force-torque, limiting the accuracy of the system due to occlusion, perspective effect, and sensory inaccuracy. Multi-modal systems have been explored to improve the robustness of automated insertion. \citet{Spector2021insertionnet, Spector2022insertionnet2} use RGB cameras and a force-torque sensor for learning contact and impedance control. \citet{chaudhury2022using} couple vision and tactile data to perform localization and pose estimation, and demonstrate that vision helps with disambiguating tactile signals for objects without distinctive features. \citet{ichiwara2022contact} leverage tactile and vision for deformable bag manipulation by performing auto-regressive prediction. \citet{hansen2022visuotactile} use a contact-gated tactile, vision and proprioceptive observation to train reinforcement learning policies. \citet{okumura2022tactile} also tackle the problem of grasp pose uncertainty for insertion by using Newtonian Variational Autoencoders to combine camera observations and tactile images. They demonstrate results for USB insertion accounting for grasp pose uncertainty in one translation direction. In this work, we separate the insertion problem into an alignment phase and an insertion phase, decoupling vision and tactile inputs and also present a novel safe self-supervised approach to data collection. We are able to handle both grasp pose rotation and translation uncertainty for the USB insertion task.

%% file: 3-problem-statement.tex
\begin{figure}[t!]
    \vspace{2pt}
    \centering
    \includegraphics[width=\linewidth]{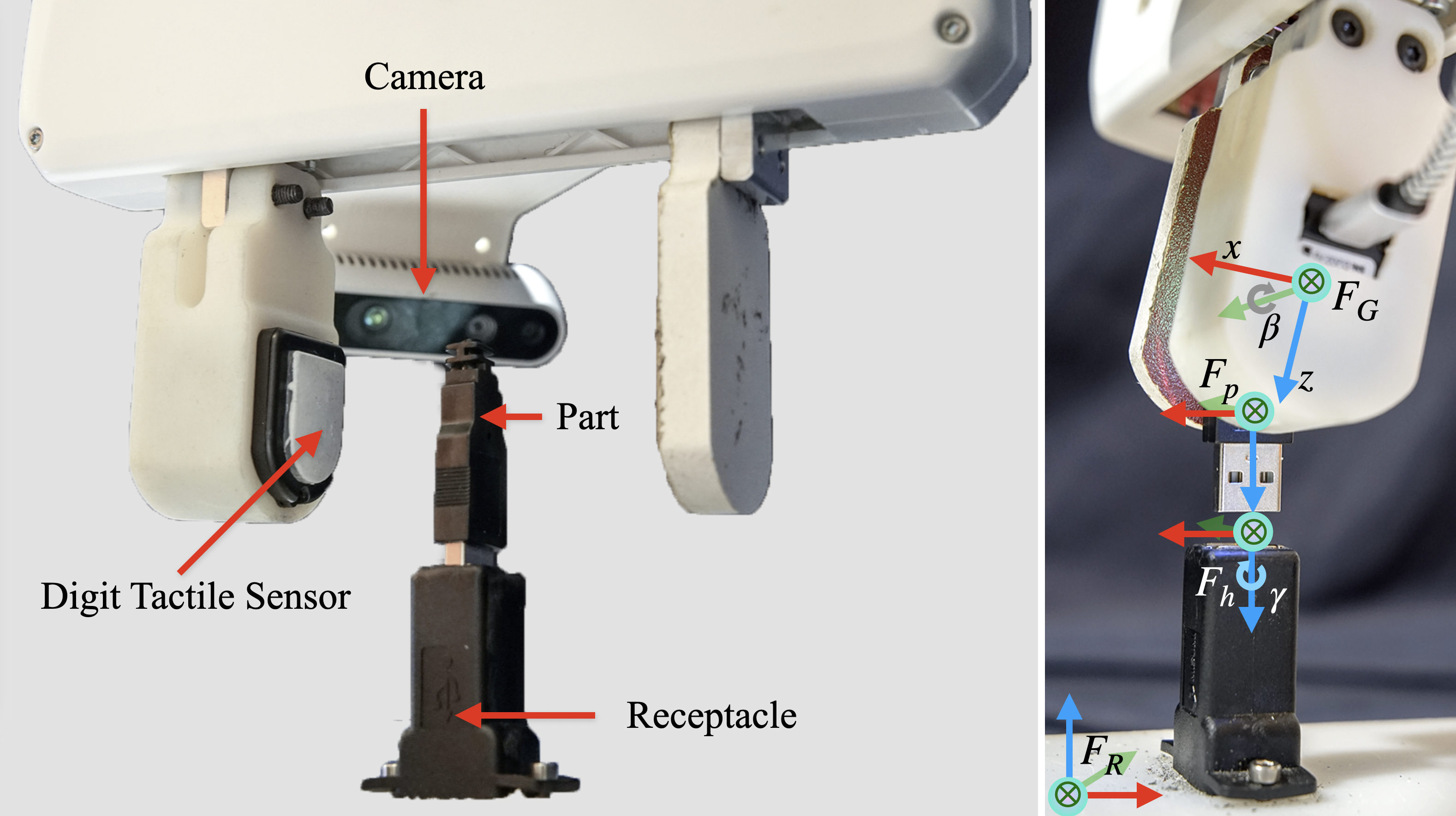}
    \caption{
    Experiment setup and coordinate system. The $x$, $y$, $z$ axes are labeled by red, green, blue respectively. We label the gripper frame, part frame, human-provided target pose frame, and robot frame as $F_G$, $F_p$, $F_h$, $F_R$ respectively. The insertion direction is defined as the $z$-axis of $F_h$. When the part is inserted, $F_h = F_p$. 
    }
    \label{fig:coor-phy}
\end{figure}
\section{Problem Statement}\label{sec:prob_form}
\textbf{Overview:}
We consider a part insertion task using a 7-DoF robot, equipped with a parallel-jaw gripper with a tactile sensor mounted on one jaw. The end-effector has a wrist-mounted RGB camera, and the robot provides reliable force-torque readings at the end-effector. 
The objective is to learn a policy that can robustly insert the part into the receptacle with an unknown part's pose within the gripper, while minimizing human inputs and part-receptacle collisions during training. Fig.~\ref{fig:coor-phy} shows the experiment setup and the coordinate frames.

\textbf{Assumptions:}
We make the following assumptions: 
\begin{enumerate}
    \item A human provides one top-down grasp pose of the part inserted in the receptacle.
    \item The robot can accurately measure force and torque either with an external sensor or internal current sensing;
    \item An experiment begins with the part pre-grasped by the robot gripper with the part grasp pose within a range relative to the human-provided pose.
    \item During data collection, the robot operates in a rectangular collision-free configuration space above the receptacle.
\end{enumerate}

\textbf{Problem Setup:} 
Given a tight-fitting receptacle for the part to be inserted into (Fig. \ref{fig:coor-phy}), we find the target insertion pose $T_{R,h}$ of the part (grasped at an unknown pose) from one human-provided imperfect demonstration. 
At any time step $t$, we have access to the RGB observation $o_{\rho}(t)$ from the wrist mounted camera, the RGB tactile image $o_{\psi}(t)$ from the DIGIT tactile sensor, and reliable force $\vec{f}(t)$ and torque $\tau(t)$ readings from the robot. Since we know the insertion axis, we parameterize the action space with 4 degrees of freedom: gripper translation in robot frame and gripper y-axis rotation. 

%% file: 4-methods.tex
\section{Method}
\label{sec:method}
\subsection{Hardware}
\begin{figure}[t]
    \centering
\begin{tikzpicture}[font=\small]
    \node[] (r1) {\includegraphics[trim=122 10 122 10, clip, height=3.2cm]{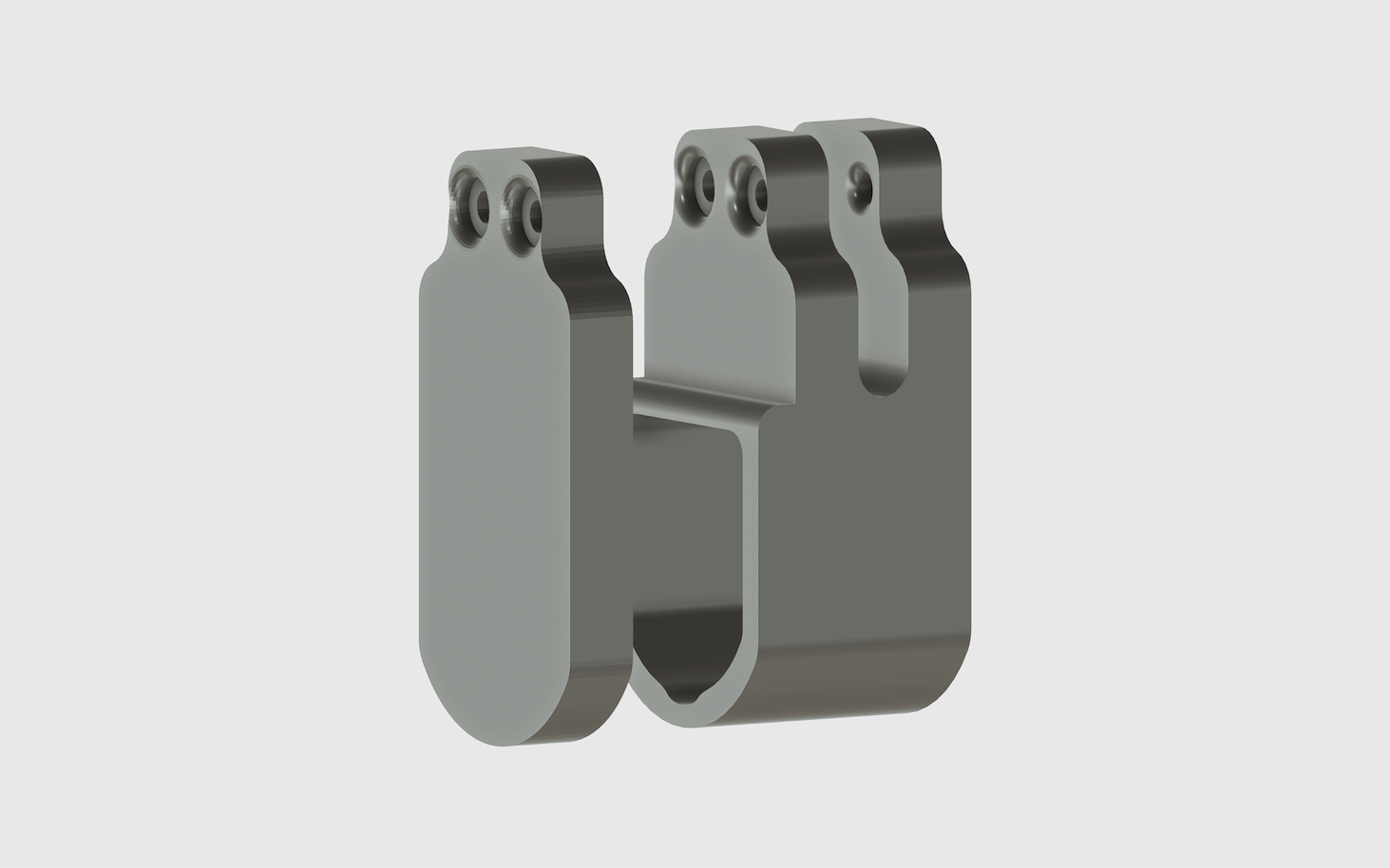}};
    \node[right=0mm of r1] (r2) {\includegraphics[trim=70 0 85 0, clip, height=3.2cm]{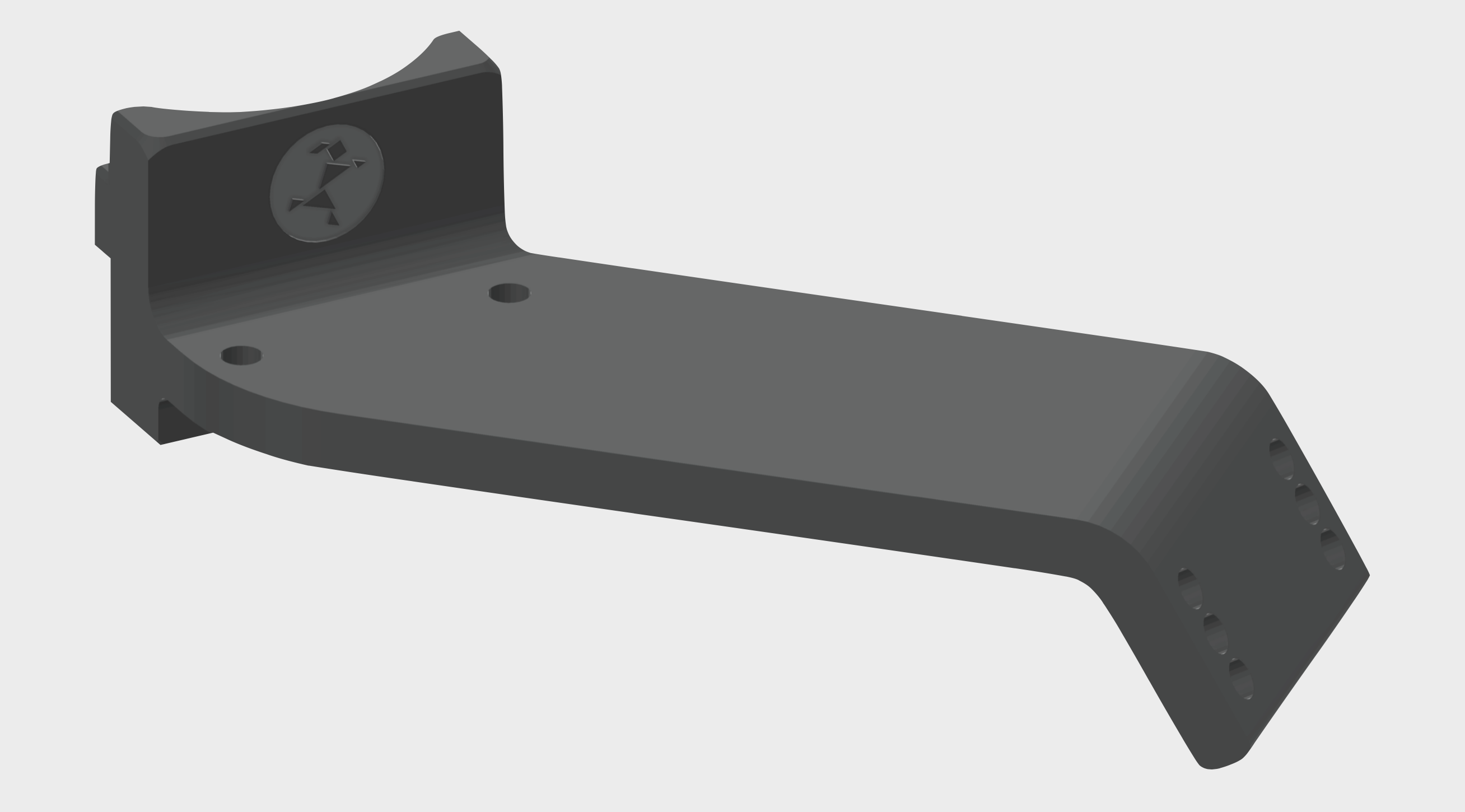}};

    \node[anchor=north, below=-1mm of r1, text width=88, align=center] {(a) Parallel Jaw Gripper};
    \node[anchor=north, below=-1mm of r2, text width=88, align=center] {(b) Camera Mount};
\end{tikzpicture}
    \caption{CAD models for the parallel jaw grippers and camera mount.}
    \label{fig:custom-hardware}
\end{figure}

We design a novel parallel jaw gripper mount to accommodate the DIGIT tactile sensor~\cite{lambeta2020digit} and camera mount (Fig.~\ref{fig:coor-phy}, ~\ref{fig:custom-hardware}). The elastomer gel on the DIGIT sensor deforms, causing torque applied to the part. This torque and the force applied from the receptacle to the part during insertion often produce undesired slippage and rotation. Therefore, we develop an asymmetric mounting setup where we mount the DIGIT sensor on one jaw with reinforcement to prevent outward bending, while keeping the other jaw flat. We apply sandpaper on the surface of the non-tactile gripper, increasing the friction to reduce slippage. We find that we can predict the part's grasp pose of a USB connector using a single DIGIT sensor.

\subsection{Self-Supervised Data Collection} \label{sec:self-supervised}

\subsubsection{One Human Provided Imperfect Target Pose}\label{data:calib}
The target pose $T_{R,h}$ is provided by a human free-driving the robot with a pre-grasped part to insert it in the receptacle from top down. Since this target pose may not have a perfect axis alignment with the receptacle, the system performs a $z$-axis alignment of the target pose. To account for the change in grasp center after axis alignment, we refine $T_{R,h}$ by finding a target pose that minimizes gripper force-torque using a grid search through a set of translations and rotations $\{T_\Delta\}$. Formally, we find
\begin{equation} \label{eq:min_ft}
    \Tilde{T}_{R,h} = \argmin_{\Tilde{T}_{R,h} \in \{T_\Delta \cdot {T}_{R,h}\}} \norm{\vec{f}(\Tilde {T}_{R,h})} + \norm{\tau(\Tilde {T}_{R,h})}.
\end{equation}
Here $\vec{f}(\Tilde {T}_{R,h})$ and $\tau(\Tilde {T}_{R,h})$ denote the 3-DoF force and torque vectors respectively when the gripper is at $\Tilde {T}_{R,h}$.
Intuitively, this objective minimizes the external force applied on the part when being unplugged, increasing the likelihood of the insertion process being reversible. 
Practically, we perform grid sampling over 5 values of $x \in [\text{-}1, 1]\text{mm}$, 5 values of $y \in [\text{-}1, 1]\text{mm}$ and 4 values of $\gamma \in [\text{-}\frac{\pi}{180}, \frac{\pi}{180}]\text{rad}$ ($x, y, \gamma$ are in $F_h$, refer to Fig.~\ref{fig:coor-phy}). The pose with minimum gripper external force-torque is recorded as the refined demonstration pose $\Tilde {T}_{R,h}$, and we have $F_h=F_G$ at $\Tilde {T}_{R,h}$. 

A cascaded impedance controller, implemented within the robot's real-time control loop, allows fine-grained force control. In case of a force violation, our system calculates a trajectory to a safe state within a single control cycle. After refinement of the target pose, we search for the minimum offset $z_{\text{min}}$ for the part to be unplugged from the receptacle. Finding the minimum height helps to determine the boundary for data collection and allows the pipeline to collect more data closer to the receptacle while reducing collisions. Iteratively, the robot moves the gripper by $-\Delta_z$ in $F_G$. We then move the gripper by $\Delta_x$ (in the $F_G$ frame) and measure $\vec{f}_x$ (x-component of the gripper force in the $F_G$ frame). If $\vec{f}_x \leq \eta$, we register the total upward distance traveled as the minimum height $z_{\text{min}}$ for removal of the part. Empirically, we find setting $\Delta_x=\Delta_z=$\SI{1}{mm}, and $\eta = \SI{3.5}{N}$ works well.
\subsubsection{Data Collection for \textbf{Alignment}}\label{data:phase1}
The part remains inserted throughout data collection. We explore grasp pose variations in 3-DoF ($x$, $z$ translation and $y$-axis rotation $\beta$) in $F_G$ (Fig.~\ref{fig:coor-phy}). We perform uniform random sampling over the range $[-3, 3]$mm, $[-8, -2]$mm, [$-\frac{\pi}{15}$, $\frac{\pi}{15}$]rad for $x, z, \beta$, with 5, 10 and 20 samples respectively. The robot closes the gripper with a  force of 70$N$ at each of the sampled poses and records the pair of tactile image readings and $x, z, \beta$. To account for the noise in the DIGIT tactile sensor, we take a median filter over 5 consecutively captured tactile images. We collect 2000 data points in 120 minutes.

\subsubsection{Data Collection for \textbf{Insertion}}\label{data:phase2}
Upon completing the tactile image collection phase, we collect robot poses and RGB images for training the insertion policy for different grasps. We perform grid sampling with 5 samples each of $x, z, \beta$ in $F_G$ within the same range as in the previous stage, resulting in a total of 125 grasps. To account for the difference between the sampled grasp $g$ and $\tilde{T}_{R,h}$, we perform minimum force-torque refinement on the sampled grasp to calculate the grasp-specific target pose $\tilde{T}_{R,h}(g)$. $\tilde{T}_{R,h}(g)$ translated by an offset of $z_{\text{min}}$ gives us the unplugged part pose $T_{\text{unplug}}(g)$.

For each grasp $g$, we collect image data for the visuoservo policy by moving the gripper to sampled points on a grid above the target pose. In particular, we uniformly sample 5 values each of $x\in[-5, 5]$mm, $y\in[-5, 5]$mm and $z\in[-5, 0]$mm in $F_R$ with respect to $T_{\text{unplug}}(g)$, resulting in 125 different translations for the gripper. For each translation, we collect \textit{one} data point that does not contain additional rotation and sample \textit{two} gripper $y$-axis rotation conditioned on $z$ to avoid collision. Specifically, given a height $z$, the two rotations are sampled from the uniform distribution $\frac{z}{5} \cdot \mathcal{U}[-\frac{\pi}{15}, 0]$rad and $\frac{z}{5} \cdot \mathcal{U}[0, \frac{\pi}{15}]$rad. These rotated data points provide the system with additional data for camera pose variation with respect to the target pose, which leads to a balanced dataset with 375 distinct gripper poses. Each data point is composed of the gripper pose (translated and rotated away from the target pose) and the corresponding RGB image observation at that pose. Upon visiting all 375 gripper poses for a given grasp, the robot moves to $T_{\text{unplug}}(g)$, performs a vertical movement to $\tilde{T}_{R,h}(g)$, and opens the gripper jaws, thereby resetting the part in the receptacle. The system repeats this data collection process for all 125 grasps without human supervision. 

\begin{figure}[t]
    \centering
    \includegraphics[width=\linewidth]{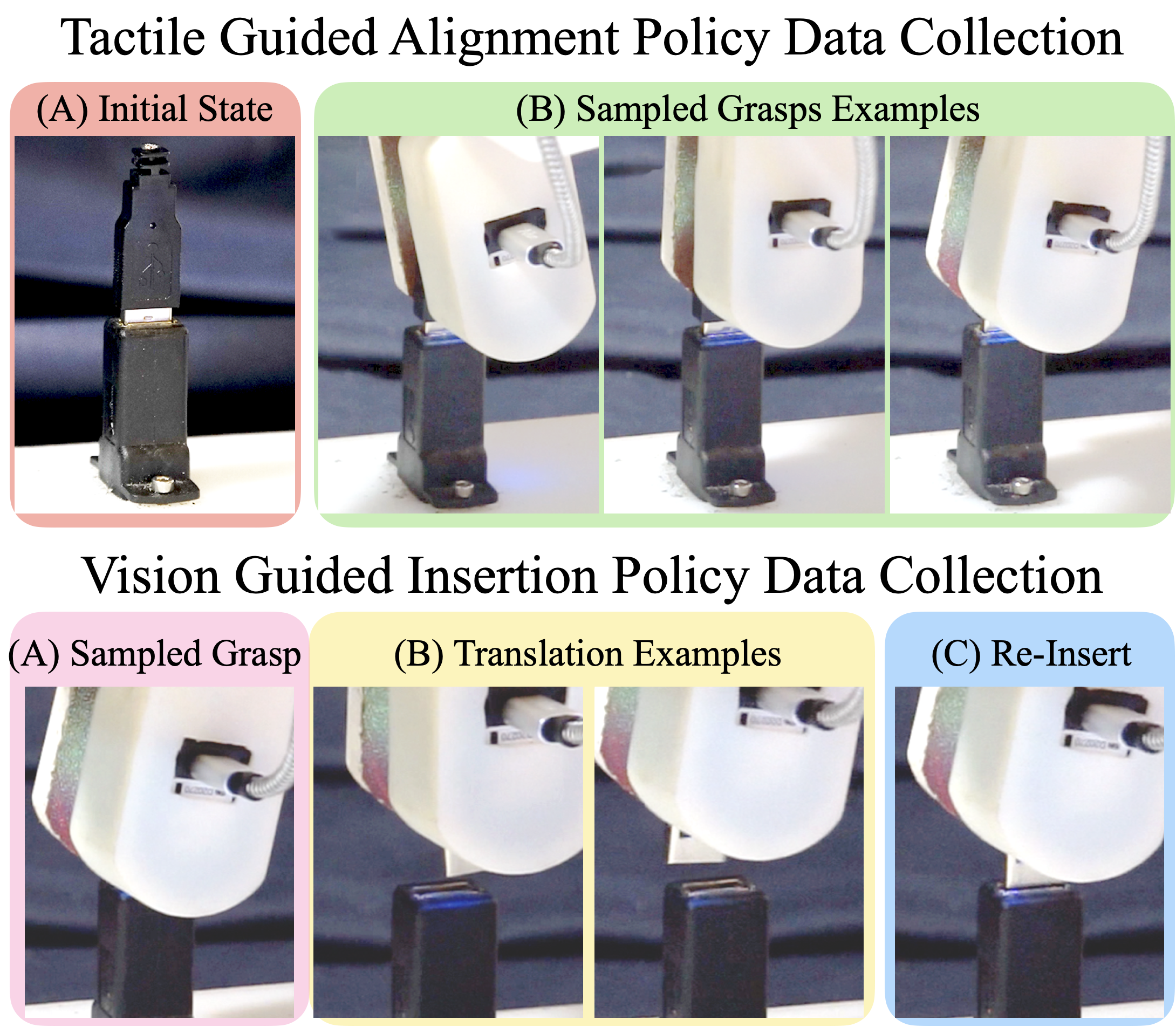}
    \caption{Data collection for alignment (Top) and insertion policies (Bottom). Data collection for the \textbf{Alignment policy} starts with the part inserted into the receptacle (Top (A)). The robot then samples and records different grasp poses and the corresponding tactile images (Top (B)). Data collection for the \textbf{Insertion policy} starts with a sampled refined grasp (Bottom (A)) and unplugs the part to apply sampled transformations (Bottom (B)). Then the robot inserts the part (Bottom (C)) and starts the next round of data collection with a different grasp pose.}
    \label{fig:datacollection}
\end{figure}
\subsection{Learning to Insert}
While human demonstrations usually serve as ``expert policies" for industrial insertion tasks, the self-supervised data collection pipeline allows us to collect ground truth actions at scale. This allows us to formulate two supervised-learning problems based on Sec.~\ref{data:phase1} and Sec.~\ref{data:phase2}. 

\subsubsection{Alignment Policy}\label{alg:align}
We use the data collected from Sec.~\ref{data:phase1} to train an alignment policy $\pi_{tac}$ that, given the tactile image, outputs the desired displacement of the gripper $T_{p,G}$ to align the part with the receptacle (Fig.~\ref{fig:splash}.B). We augment the tactile images by randomly jittering the brightness and contrast over the range $\mathcal{U}[0.7, 1.3]$. 

\subsubsection{Insertion Policy}\label{alg:insert}
We use data collected from Sec.~\ref{data:phase2} to train a visuoservo insertion policy $\pi_{vis}$ taking normalized camera observation, gripper $y$-axis rotation $\beta$ and $x, y$ translations in $F_h$ as inputs, and predict the action: desired translation $\Delta_x, \Delta_y$ and rotation $\Delta_\beta$ in $F_R$ (Fig.~\ref{fig:splash}.C). We augment camera observations by randomly jittering the brightness and contrast over the range $\mathcal{U}[0.7, 1.3]$. 

We use RegNet 3.2GF~\cite{radosavovic2020designing} as the backbone for both policies. For the alignment policy, we replace the last layer of RegNet with a linear layer with 3 outputs. For the insertion policy, we concatenate the robot's pose with the latent vector of the image and replace the last layer with a linear layer with 3 outputs. For both networks, we use a batch size of 64, a learning rate of 1e-3, and a learning rate decay of 0.99 for every 100 gradient steps. We pick the mean squared error as the loss function and use the Adam optimizer~\cite{kingma2014adam}.

\subsection{Execution of Insertion}
\input{4.2.execution.tex}
To avoid catastrophic failure (i.e. collision between the part and the surrounding environment or moving out of the training set distribution), we deliberately formulate the visuoservo policy to only control $x,y$-axis translation but not $z$-axis translation since the insertion direction is already aligned with the $z$-axis by the {\em Align} policy.  The formulation is detailed in Algorithm ~\ref{alg:alg}. The execution procedure starts by inferring the grasp pose from the tactile image via $\pi_{tac}$ (line 2). The system then performs the insertion axis alignment of the part with the receptacle (line 3). We measure the z-direction force based on the gripper force-torque sensor (line 12). We then calculate rotation and translation based on the camera observation via $\pi_{vis}$ (line 5, 10) and execute the corresponding actions (line 8) until the action has a norm smaller than $\epsilon$ (line 7). Note that the rotation prediction from the \textit{Insertion Policy} is not used, because the part is aligned with the insertion axis by the \textit{Alignment Policy}. When the action converges, we lower the gripper in the $z$ direction by a step size of $d_z$ (line 11) and continue to query $\pi_{vis}$ until the force constraint is satisfied or the number of attempts exceeds the horizon $H$ (line 13-14). We empirically set $d_z=1.5$mm. Empirically, we find setting the action norm $\epsilon = 0.0005$, $d_z=1.5$mm and $H=200$ works well, as the force-based termination condition is usually triggered first for a successful or unsuccessful insertion.

%% file: 4.2.execution.tex
\begin{algorithm}[t]
\vspace{2pt}
\SetAlgoLined
    \textbf{Input:} Tactile Image $o_{\rho}(t)$, Camera Image $o_{\psi}(t)$, tactile based grasp pose estimation network $\pi_{tac}$, and visuoservo insertion policy $\pi_{vis}$, Target Pose $\tilde{T}_{R,h}$, Minimum Wrench Height $z_{\text{min}}$, action norm threshold $\epsilon$, z direction step size $d_z$ \\
    $T_{p,G} = \pi_{tac}(o_{\rho}(t))$ \\
    Move gripper to $\tilde{T}_{R,h} T_{p,G} + 2z_{\text{min}}$\\
    attempts = 0\\
    $(\Delta_x,\Delta_y,\Delta_\beta) = \pi_{vis}(o_{\psi}(t), T_{R,G}(t))$\\
    \While{True}{
        \While{$\norm{[\Delta_x,\Delta_y]} > \epsilon$ and attempts $<$ H}{
            Move gripper by $(\Delta_x,\Delta_y)$\\
            attempts = attempts + 1\\
            $(\Delta_x,\Delta_y,\Delta_\beta) = \pi_{vis}(o_{\psi}(t), T_{R,G}(t))$\\
        }
        Translate gripper in insertion direction by $d_z$\\
        Measure gripper force-torque in z-axis as $F_z$\\
        \uIf{$F_z > 15$N or attempts $\geq$ H} {
            \textbf{Terminate}
        }
    }
    \caption{Policy Execution Procedure}
    \label{alg:alg}
\end{algorithm}

%% file: 5-experiment-and-result.tex
\vspace{-5pt}
\section{Experiments}
\vspace{-5pt}
\label{sec:experiments_and_results}

\begin{table}[t!]
\vspace{5pt}
\centering
\begin{tabular}{lp{1cm}p{1cm}p{2.5cm}}
\hline
Algorithms    & IL   & TD3      & Proposed Approach \\ \hline
Success/Total & 1/45 & 0/45 & 45/45            \\ \hline
\end{tabular}
\caption{Results suggest that (1) the IL trained on 50 human demonstrations is insufficient for training an accurate part pose estimation model, and (2) frequent slippage and rotations of the USB caused by collisions with the receptacle lead to failure in training TD3. Our approach outperforms both baseline policies.}
\label{tbl:main}
\end{table}
\subsection{Experiment Setup}
\vspace{-5pt}
We focus on the USB insertion task on the NIST taskboard. We use a 7-DoF Franka Emika robot with a parallel gripper, where one DIGIT tactile sensor is mounted on the inside of one of the fingers. An Intel RealSense camera is mounted offset from the gripper (Fig.~\ref{fig:custom-hardware}). To control the Cartesian position of the Franka robot, a time-optimal trajectory respecting velocity, acceleration, and jerk constraints is applied to the policy's positional output \cite{berscheid2021jerk}. We use grid sampling to obtain 5 values of $\beta$ ranging from $[-\frac{\pi}{20}, \frac{\pi}{20}]$rad, 3 translations in x and z from the range $[-3, 3]$mm and $[-6, -2]$mm in $F_h$, resulting in a test set of 45 different grasp configurations that lie in the training distribution of the algorithms.

\subsection{Experimental Procedure}
\vspace{-5pt}
At the beginning of each test experiment, the USB connector (part) is pre-grasped by the robot with a grasp pose selected from the test set and located at a position with a $z$-axis translation of $2z_{\text{min}}$ relative to $F_h$. At this starting pose, the gripper is aligned vertically down while the USB connector is misaligned with the receptacle in both translation and rotation as in Fig.~\ref{fig:splash}(A). 
The robot first executes the {\em Align} policy to estimate the part grasp pose and aligns it with the insertion direction of the receptacle as in  Fig.~\ref{fig:splash}(B). It then uses the {\em Insert} policy to visuoservo and inserts the part into the receptacle as in Fig.~\ref{fig:splash}(C-D). The robot then resets to the next grasp by releasing the part, re-grasping it, raising it to a start pose as outlined above, and executing the {\em Align} and {\em Insert} policies for this new grasp.  It steps through all the test grasp poses using the same procedure. 

An experiment terminates if the gripper frame ($F_G$) force in the $z$ direction $\vec{f}_z(t)$ exceeds \SI{15}{N}. An experiment trial is considered successful if the gripper is within \SI{5}{mm} of the target pose in $H=200$ iterations, upon which we also visually inspect whether the insertion is successful. In this set of experiments, the refined human-provided target pose $\tilde{T}_{R,h}$ is provided as an input to the policies. In Sec.~\ref{sec:ablation2}, we perform ablation studies on noisy target poses.

\subsection{Comparison}
\vspace{-5pt}
The method is designed with two objectives: (1) minimizing human intervention so that ideally no human needs to be involved in data collection or training of the policy, and (2) minimizing collision among the robot, the part, and the environment. We compare our approach with two baseline learning methods described below with the same environment setup (using the same grippers and camera mount). 
\subsubsection{Twin Delayed Deep Deterministic Policy Gradient(TD3)} An off-policy, online reinforcement learning policy~\cite{fujimoto2018addressing} that learns the end-to-end part insertion. This baseline satisfies objective (1) but violates objective (2)  --- i.e.\ it is incapable of avoiding collisions among the robot, the part, and the environment. We simplify the problem to a fixed, axis-aligned grasp pose and restrict the action space to translations only, so that the policy only have to learn insertion instead of both alignment and insertion. The learned policy runs with a frequency of \SI{10}{Hz}. Since the policy outputs Cartesian position changes, we controls the Cartesian velocity of the robot, which is equivalent to Cartesian position changes per time step. Due to the part's axis-alignment with the receptacle, the policy's input can be restricted to the (low-dimensional) robot pose, velocity, and the force-torque at frame $F_G$.
We use the default TD3 implementation of Ray RLLib~\cite{moritz2018ray}. If the force applied on the gripper exceeds \SI{15}{N}, the episode terminates and the robot resets to a safe starting pose.
\subsubsection{Imitation learning (IL)} Imitation learning from 50 human demonstrations of insertion trained with behavior cloning. Each human demonstration starts with a randomly sampled grasp pose as in Sec.~\ref{data:phase1}. This baseline algorithm violates objective (1) but satisfies objective (2), where the human demonstrator selects actions that minimize collision between the part and the environment. It takes about 30 minutes to provide all these human demonstrations. 

\begin{table}[t]
\vspace{5pt}
\centering
\begin{tabular}{lp{1.3cm}p{1.3cm}p{1.3cm}}
\hline
{}   & $x$ (mm)   & $z$ (mm)     & $\beta$ (rad) \\ \hline
Mean Error & $8.97$e-$2$ & $1.46$e-$1$ & $5.59$e-$3$            \\
Standard Deviation & $4.89$e-$3$ & $6.62$e-$2$ & $4.89$e-$3$            \\ \hline
\end{tabular}
\caption{Mean and standard deviation of the error in predicting part pose ($x, z, \beta$) by the tactile-based alignment policy on the test set of 45 grasps.}
\label{tbl:tac_error}
\end{table}

\subsection{Results}
\vspace{-5pt}
The results are summarized in Table.~\ref{tbl:main}. The imitation learning agent (IL) is only able to perform a single successful insertion out of the 45 grasp poses. Intuitively, 50 different grasp configurations from human demonstrations are not sufficient for training an accurate part pose estimation model; additionally, for different grasp poses, at the same gripper pose, two different human demonstrations may exist. The multi-modality in the distribution of target insertion poses contributes to the failure of the IL policy. Training the TD3 policy in the physical environment led to divergent results in all 5 training trials we attempted. In all cases, the part collides with the receptacle, leading to a drastic change in the grasp pose. This cannot be corrected directly since there is no reset procedure that can systematically recover the gripper to the original state without human supervision. Our approach succeeds for every single grasp pose tested. Empirically, we find that the part rotation and translation predicted from tactile images are fairly accurate (refer to Table.~\ref{tbl:tac_error}). 

%% file: 6-ablation.tex
\vspace{-5pt}
\subsection{Ablation Studies}\label{sec:ablation}
\subsubsection{Effects of Leveraging Force-Torque Sensing in Data Collection} \label{sec:ablation1}
We compare the completion rate of the data collection process for insertion with or without grasp pose refinement. We consider the following three different methods for refining the target pose gained from the human demonstration $T_{R,h}$: 1) \textbf{ZA}: apply $z$-axis alignment on $T_{R,h}$, 2) \textbf{ZAWF}: perform minimum force-torque refinement only once for $T_{R,h}$ after $z$-axis alignment (this step is only performed for the first grasp and the results reused for all grasps) and 3) \textbf{ZAWFG}: perform $z$-axis alignment for $T_{R,h}$ and apply minimum force-torque refinement separately for each grasp. 

At the beginning of each experiment, a human provides $T_{R,h}$ by free-driving the robot with one pre-grasped part to insert the part. A total of 125 different grasp poses are sampled. For each grasp pose $T_{h,G}$, we calculate the grasp pose in robot frame by $T_{R,G} = \tilde{T}_{R,h}T_{h,G}$ with $\tilde{T}_{R,h}$ determined by one of the three methods \textbf{ZA}, \textbf{ZAWF} or \textbf{ZAWFG}. The robot grasps the part with the pose $T_{R,G}$, lifts the part, and tries to re-insert the part. If insertion is successful, the robot executes the next grasp otherwise the experiment terminates. We report total number of successful insertions before termination (Table.~\ref{tbl:min_ft}). We repeat the experiment three times for each method with different human demonstrations. 

After applying $z$-axis alignment for the human-provided target pose (\textbf{ZA}), the insertion fails as the center of the grasp is not aligned with the center of the receptacle. \textbf{ZAWF} addresses this issue by using minimum pose refinement, and can perform successful insertions. However, since the pose refinement is specific to the human demonstration, the refinement is not sufficiently granular, leading to failures when the new grasp configuration has a large $y$-axis rotation. \textbf{ZAWFG} performs pose refinement for each of the grasps, resulting in consistent insertion performance. \textbf{ZAWFG} needs to wait for the force measurements to settle and thus takes longer to execute. 

\begin{table}[t!]
\vspace{5pt}
\begin{tabular}{ccccc}
\hline
Setup   & Trial 1 & Trial 2 & Trial 3 & Mean$\pm$Standard Error \\ \hline
ZA      & 0/125       & 0/125       & 0/125       & 0.0$\pm$0.0$\%$                \\
ZAWF   & 125/125     & 125/125     & 34/125      & 75.7$\pm$19.8$\%$              \\
ZAWFG & 125/125     & 125/125     & 125/125     & 100.0$\pm$0.0$\%$              \\
\hline
\end{tabular}
\caption{Comparing data collection success rate. We measure the number of successful insertions until failure for 125 different grasps configurations. We compare Human Demonstration with axis alignment (\textbf{ZA}), Single Minimum Force-Torque Refinement (\textbf{ZAWF}), and Minimum Force-Torque Refinement for all grasps (\textbf{ZAWFG}). We report the mean success rate and the standard error for three distinct human-provided target poses.}
\label{tbl:min_ft}
\end{table}

\subsubsection{Exploring Utility of Tactile and Vision  Information}\label{sec:ablation2}
We perform study the relative benefits of using tactile and vision for insertion tasks. We test 3 different approaches: 
(1) A Tactile Only approach (2) A Vision Only approach trained using a limited amount of the camera observation data and (3) a Combined Approach. This ablation study differs from our earlier experiments by injecting a uniformly sampled noise in the range $\pm 1$mm into the target pose's $x, y$ translation to imitate imprecise knowledge of the target pose. 

The Tactile Only approach attempts the entire insertion task in a single phase using the tactile information to align the USB connector with the receptacle and then move straight down to insert it. For the purpose of this study, we train a new Vision Only model using a third of the collected camera observation data set that have no additional gripper rotation.
This mimics a mono-view visuoservo model for receptacle localization and top-down insertion. We modify the insertion motion from Sec.~\ref{alg:insert} so the model only uses camera observation and the $y$-axis rotation of the gripper, and it only outputs translation in $x$ and $y$ based on the same regression objective as in Alg.~\ref{alg:insert}. The combined approach sequences a Tactile Only approach in an \textit{Align} phase with the modified Vision Only approach in an \textit{Insert} phase. We perform experiments with the three different approaches with the same test set as in Sec.~\ref{sec:experiments_and_results} and report results in Table~\ref{tbl:ablation2}.

With noisy target pose, the Tactile Only model succeeds only 21/45 times since the model does not estimate the target receptacle state. Since the Vision Only model is constrained to translation actions and is trained on a limited set of data that does not include additional gripper rotation, it inserts the part when the grasps do not have any rotation (8/9 successes) but fails otherwise (for a total of 8/45 successes). A separate Vision Only model trained with all the data is similarly unsuccessful (11/45 successes), indicating the importance of the ability to correct for grasp pose variation using tactile data. The combination of the two models outperforms either model by leveraging the part rotation prediction (using tactile) and implicitly estimating the environment state (using visual information), suggesting that tactile and vision observation jointly reduce the uncertainties in the insertion problem.
\begin{table}[t!]
\vspace{5pt}
\centering
\begin{tabular}{lp{1.8cm}p{1.8cm}p{1.8cm}p{1.8cm}p{1.8cm}}
\hline
Algorithm & Tactile Only & Vision Only (No Rot) & Tactile + Vision (No Rot) \\ \hline
Success/Total & 21/45 & 8/45 & 40/45 \\
\hline
\end{tabular}
\caption{Ablation study with noisy target poses comparing single-phase Tactile Only, modified Vision Only, and a Combined two-phase approach leveraging tactile and visual information.}
\label{tbl:ablation2}
\end{table}

%% file: 7-summary.tex
\vspace{-5pt}
\section{Limitations and Conclusion}
\vspace{-5pt}
Despite promising results, this method has not been tested for generalization to other types of assembly tasks, objects with more complex geometries, or objects that are larger than the tactile sensor. Parts made of different materials may require distinct maximal forces; the grid search for finetuning the insertion pose lengthens the data collection process. The robot must also unplug the part, which can pose a challenge as some parts are designed to be difficult to remove (i.e. an Ethernet connector). This work did not measure the time required for data collection. Future research can improve on the time required for collecting data. In summary, we present a safe, self-supervised method for learning a visuo-tactile insertion policy in real industrial settings with unknown grasp poses. We achieve this by using force-torque sensing to refine human-demonstrated target poses and constructing a two-phase approach to insertion that separates the task into alignment and insertion based on tactile and visual feedback. 

%% file: references.bib
@STRING{icra = {{Proc. {IEEE} Int. Conf. Robotics and Automation (ICRA)}}}

@STRING{iros = {Proc. IEEE/RSJ Int. Conf. on Intelligent Robots and Systems (IROS)}}

@STRING{rss = {Proc. Robotics: Science and Systems (RSS)}}

@STRING{corl = {Conf. on Robot Learning (CoRL)}}

@article{Wen2022once,
  author    = {Bowen Wen and
               Wenzhao Lian and
               Kostas E. Bekris and
               Stefan Schaal},
  title     = {You Only Demonstrate Once: Category-Level Manipulation from Single
               Visual Demonstration},
  journal   = {Robotics: Science and Systems (RSS)},
  year      = {2022}
}

@article{Lian2021benchmark,
  author    = {Wenzhao Lian and
               Tim Kelch and
               Dirk Holz and
               Adam Norton and
               Stefan Schaal},
  title     = {Benchmarking Off-The-Shelf Solutions to Robotic Assembly Tasks},
  journal   = {Proc. {IEEE/RSJ} Int. Conf. on Intelligent Robots and Systems (IROS)},
  year      = {2021}
}

@article{Spector2022insertionnet2,
  title={InsertionNet 2.0: Minimal Contact Multi-Step Insertion Using Multimodal Multiview Sensory Input},
  author={Oren Spector and Vladimir Tchuiev and Dotan Di Castro},
  journal={Proc. {IEEE} Int. Conf. Robotics and Automation (ICRA)},
  year={2022}
}

@article{Spector2021insertionnet,
  author    = {Oren Spector and
               Dotan Di Castro},
  title     = {InsertionNet - {A} Scalable Solution for Insertion},
  journal   = {{IEEE} Robotics and Automation Letters},
  year      = {2021}
}

@article{Florence2021ibc,
  author    = {Pete Florence and
               Corey Lynch and
               Andy Zeng and
               Oscar Ramirez and
               Ayzaan Wahid and
               Laura Downs and
               Adrian Wong and
               Johnny Lee and
               Igor Mordatch and
               Jonathan Tompson},
  title     = {Implicit Behavioral Cloning},
  journal   = {Conf. on Robot Learning (CoRL)},
  year      = {2021}
}

@article{Li2014gelsight,
  title={Localization and manipulation of small parts using GelSight tactile sensing},
  author={Rui Li and Robert W. Platt and Wenzhen Yuan and Andreas ten Pas and Nathan Roscup and Mandayam A. Srinivasan and Edward H. Adelson},
  journal={Proc. {IEEE/RSJ} Int. Conf. on Intelligent Robots and Systems (IROS)},
  year={2014}
}

@article{Yuan2017gelsight,
  title={Gelsight: Highresolution robot tactile sensors for estimating geometry and force},
  author={Wenzhen Yuan and Siyuan Dong and Edward H Adelson},
  journal={Sensors 17},
  year={2017}
}

@article{lambeta2020digit,
  title={Digit: A novel design for a low-cost compact high-resolution tactile sensor with application to in-hand manipulation},
  author={Lambeta, Mike and Chou, Po-Wei and Tian, Stephen and Yang, Brian and Maloon, Benjamin and Most, Victoria Rose and Stroud, Dave and Santos, Raymond and Byagowi, Ahmad and Kammerer, Gregg and others},
  journal={IEEE Robotics and Automation Letters},
  year={2020},
}

@article{Kelestemur2022tactilepose,
  title={Tactile Pose Estimation and Policy Learning for Unknown Object Manipulation},
  author={Tarik Kelestemur and Robert Platt and Taskin Padir},
  journal={Int. Conf. on Autonomous Agents and Multiagent Systems (AAMAS)},
  year={2022},
}

@article{chaudhury2022using,
  title={Using Collocated Vision and Tactile Sensors for Visual Servoing and Localization},
  author={Chaudhury, Arkadeep Narayan and Man, Timothy and Yuan, Wenzhen and Atkeson, Christopher G},
  journal={IEEE Robotics and Automation Letters},
  volume={7},
  number={2},
  pages={3427--3434},
  year={2022},
  publisher={IEEE}
}

@inproceedings{ichiwara2022contact,
  title={Contact-rich manipulation of a flexible object based on deep predictive learning using vision and tactility},
  author={Ichiwara, Hideyuki and Ito, Hiroshi and Yamamoto, Kenjiro and Mori, Hiroki and Ogata, Tetsuya},
  booktitle={2022 International Conference on Robotics and Automation (ICRA)},
  pages={5375--5381},
  year={2022},
  organization={IEEE}
}

@inproceedings{hansen2022visuotactile,
  title={Visuotactile-RL: Learning Multimodal Manipulation Policies with Deep Reinforcement Learning},
  author={Hansen, Johanna and Hogan, Francois and Rivkin, Dmitriy and Meger, David and Jenkin, Michael and Dudek, Gregory},
  booktitle={2022 International Conference on Robotics and Automation (ICRA)},
  pages={8298--8304},
  year={2022},
  organization={IEEE}
}

@article{NIST,
author={K. Kimble and K. Van Wyk and J. Falco and E. Messina and Y. Sun and M. Shibata and W. Uemura and Y. Yokokohji}, 
title={Benchmarking protocols for evaluating small parts robotic assembly systems},
booktitle={IEEE Robotics and Automation Letters},
volume={5(2)},
pages={883–889}, 
year={2020},
organization={IEEE}
}

@inproceedings{kingma2014adam,
  title={Adam: A method for stochastic optimization},
  author={Kingma, Diederik P and Ba, Jimmy},
  journal={Proceedings of the 3rd International Conference on Learning Representations (ICLR)},
  year={2015}
}

@inproceedings{fujimoto2018addressing,
  title={Addressing function approximation error in actor-critic methods},
  author={Fujimoto, Scott and Hoof, Herke and Meger, David},
  booktitle={International conference on machine learning},
  pages={1587--1596},
  year={2018},
  organization={PMLR}
}

@article{okumura2022tactile,
  title={Tactile-Sensitive NewtonianVAE for High-Accuracy Industrial Connector-Socket Insertion},
  author={Okumura, Ryo and Nishio, Nobuki and Taniguchi, Tadahiro},
  journal={arXiv preprint arXiv:2203.05955},
  year={2022}
}

@inproceedings{wang2021elastic,
  title={Elastic tactile simulation towards tactile-visual perception},
  author={Wang, Yikai and Huang, Wenbing and Fang, Bin and Sun, Fuchun and Li, Chang},
  booktitle={Proceedings of the 29th ACM International Conference on Multimedia},
  pages={2690--2698},
  year={2021}
}

@inproceedings{radosavovic2020designing,
  title={Designing network design spaces},
  author={Radosavovic, Ilija and Kosaraju, Raj Prateek and Girshick, Ross and He, Kaiming and Doll{\'a}r, Piotr},
  booktitle={Proceedings of the IEEE/CVF conference on computer vision and pattern recognition},
  pages={10428--10436},
  year={2020}
}

@article{narang2022factory,
  title={Factory: Fast Contact for Robotic Assembly},
  author={Narang, Yashraj and Storey, Kier and Akinola, Iretiayo and Macklin, Miles and Reist, Philipp and Wawrzyniak, Lukasz and Guo, Yunrong and Moravanszky, Adam and State, Gavriel and Lu, Michelle and others},
  journal={Robotics: Science and Systems (RSS)},
  year={2022}
}

@Article{Wang2022TACTO,
  author   = {Wang, Shaoxiong and Lambeta, Mike and Chou, Po-Wei and Calandra, Roberto},
  title    = {{TACTO}: A Fast, Flexible, and Open-source Simulator for High-resolution Vision-based Tactile Sensors},
  journal  = {IEEE Robotics and Automation Letters (RA-L)},
  year     = {2022},
  volume   = {7},
  number   = {2},
  pages    = {3930--3937},
  issn     = {2377-3766},
  doi      = {10.1109/LRA.2022.3146945},
  url      = {https://arxiv.org/abs/2012.08456},
}

@inproceedings{zhao2022offline,
  title={Offline meta-reinforcement learning for industrial insertion},
  author={Zhao, Tony Z and Luo, Jianlan and Sushkov, Oleg and Pevceviciute, Rugile and Heess, Nicolas and Scholz, Jon and Schaal, Stefan and Levine, Sergey},
  booktitle={2022 International Conference on Robotics and Automation (ICRA)},
  pages={6386--6393},
  year={2022},
  organization={IEEE}
}

@inproceedings{schoettler2020meta,
  title={Meta-reinforcement learning for robotic industrial insertion tasks},
  author={Schoettler, Gerrit and Nair, Ashvin and Ojea, Juan Aparicio and Levine, Sergey and Solowjow, Eugen},
  booktitle={2020 IEEE/RSJ International Conference on Intelligent Robots and Systems (IROS)},
  pages={9728--9735},
  year={2020},
  organization={IEEE}
}

@inproceedings{schoettler2020deep,
  title={Deep reinforcement learning for industrial insertion tasks with visual inputs and natural rewards},
  author={Schoettler, Gerrit and Nair, Ashvin and Luo, Jianlan and Bahl, Shikhar and Ojea, Juan Aparicio and Solowjow, Eugen and Levine, Sergey},
  booktitle={2020 IEEE/RSJ International Conference on Intelligent Robots and Systems (IROS)},
  pages={5548--5555},
  year={2020},
  organization={IEEE}
}

@inproceedings{johannink2019residual,
  title={Residual reinforcement learning for robot control},
  author={Johannink, Tobias and Bahl, Shikhar and Nair, Ashvin and Luo, Jianlan and Kumar, Avinash and Loskyll, Matthias and Ojea, Juan Aparicio and Solowjow, Eugen and Levine, Sergey},
  booktitle={2019 International Conference on Robotics and Automation (ICRA)},
  pages={6023--6029},
  year={2019},
  organization={IEEE}
}

@article{luo2021robust,
  title={Robust multi-modal policies for industrial assembly via reinforcement learning and demonstrations: A large-scale study},
  author={Luo, Jianlan and Sushkov, Oleg and Pevceviciute, Rugile and Lian, Wenzhao and Su, Chang and Vecerik, Mel and Ye, Ning and Schaal, Stefan and Scholz, Jon},
  journal={Robotics: Science and Systems (RSS)},
  year={2021}
}

@article{berscheid2021jerk,
  title={Jerk-limited Real-time Trajectory Generation with Arbitrary Target States},
  author={Berscheid, Lars and Kr{\"o}ger, Torsten},
  journal={Robotics: Science and Systems XVII},
  year={2021}
}

@article{lozano1984automatic,
  title={Automatic synthesis of fine-motion strategies for robots},
  author={Lozano-Perez, Tomas and Mason, Matthew T and Taylor, Russell H},
  journal={The International Journal of Robotics Research},
  volume={3},
  number={1},
  pages={3--24},
  year={1984},
  publisher={Sage Publications Sage CA: Thousand Oaks, CA}
}

@article{lozano1986motion,
  title={Motion planning and the design of orienting devices for vibratory part feeders},
  author={Lozano-P{\'e}rez, Tomas},
  journal={in IEEE Journal Of Robotics And Automation. MIT AI Laboratory},
  year={1986}
}

@article{goldberg1993orienting,
  title={Orienting polygonal parts without sensors},
  author={Goldberg, Kenneth Y},
  journal={Algorithmica},
  volume={10},
  number={2},
  pages={201--225},
  year={1993},
  publisher={Springer}
}

@article{natarajan1989some,
  title={Some paradigms for the automated design of parts feeders},
  author={Natarajan, Balas K},
  journal={The International journal of robotics research},
  volume={8},
  number={6},
  pages={98--109},
  year={1989},
  publisher={Sage Publications Sage CA: Thousand Oaks, CA}
}

@article{mckee1985automatic,
  title={Automatic Assembly by G. Boothroyd, C. Poli and LE Murch, Marcel Dekker, New York, 378 pp., 1982},
  author={McKee, KE},
  journal={Robotica},
  volume={3},
  number={3},
  pages={195--196},
  year={1985},
  publisher={Cambridge University Press}
}

@inproceedings{de1989correct,
  title={A correct and complete algorithm for the generation of mechanical assembly sequences},
  author={De Mello, LS Homem and Sanderson, Arthur C},
  booktitle={1989 IEEE International Conference on Robotics and Automation},
  pages={56--57},
  year={1989},
  organization={IEEE Computer Society}
}

@article{koga2022cad,
  title={On CAD Informed Adaptive Robotic Assembly},
  author={Koga, Yotto and Kerrick, Heather and Chitta, Sachin},
  journal={arXiv preprint arXiv:2208.01773},
  year={2022}
}

@inproceedings{moritz2018ray,
  title={Ray: A distributed framework for emerging AI applications},
  author={Moritz, Philipp and Nishihara, Robert and Wang, Stephanie and Tumanov, Alexey and Liaw, Richard and Liang, Eric and Elibol, Melih and Yang, Zongheng and Paul, William and Jordan, Michael I and others},
  booktitle={13th USENIX Symposium on Operating Systems Design and Implementation (OSDI 18)},
  pages={561--577},
  year={2018}
}

@article{qiao1995fine,
  title={Fine motion strategies for robotic peg-hole insertion},
  author={Qiao, Hong and Dalay, BS and Parkin, RM},
  journal={Proceedings of the Institution of Mechanical Engineers, Part C: Journal of Mechanical Engineering Science},
  volume={209},
  number={6},
  pages={429--448},
  year={1995},
  publisher={SAGE Publications Sage UK: London, England}
}

@article{Fan2019framework,
  title={A Learning Framework for High Precision Industrial Assembly},
  author={Yongxiang Fan and Jieliang Luo and Masayoshi Tomizuka},
  journal={2019 International Conference on Robotics and Automation (ICRA)},
  year={2019},
  pages={811-817}
}
